\documentclass[9pt,conference]{IEEEtran}
\usepackage{dcase2025}

\usepackage{bm} % for bold math symbols (incl. Greek letters) with \bm{}

\usepackage{acronym}
\newacro{SNR}{Signal-to-Noise Ratio}
\newacro{SSL}{Self Supervised Learning}
\newacro{T-WA}{Time-Weighted Averaging}
\newacro{T-A}{Time-Averaged}
\newacro{mAP}{Mean Average Precision}
\usepackage{graphicx}  % For resizing tables
\usepackage{booktabs}  % For better table lines
\usepackage{makecell}
\usepackage{float}
\usepackage{multirow}

\usepackage{dcase2025,amsmath,graphicx,url,times,booktabs, tabularx}

\title{Crossing the Species Divide:\\Transfer Learning from Speech to Animal Sounds}

\name{Jules Cauzinille$^{1, 4}$,
      Marius Miron$^{2}$,
      Olivier Pietquin$^{2}$,
      Masato Hagiwara$^{2}$,
      Ricard Marxer$^{3, 4}$,
      Arnaud Rey$^{2, 4}$,
      Benoit Favre$^{1, 4}$}
\address{$^{1}$Aix-Marseille University, France \;
$^{2}$Earth Species Project, USA\\
$^{3}$Université de Toulon, France \;
$^{4}$ILCB, France
}

\begin{document}

\maketitle

\begin{abstract}

Self-supervised speech models have demonstrated impressive performance in speech processing, but their effectiveness on non-speech data remains underexplored. We study the transfer learning capabilities of such models on bioacoustic detection and classification tasks. We show that models such as HuBERT, WavLM, and XEUS can generate rich latent representations of animal sounds across taxa. We analyze the models properties with linear probing on time-averaged representations. We then extend the approach to account for the effect of time-wise information with other downstream architectures. Finally, we study the implication of frequency range and noise on performance. Notably, our results are competitive with fine-tuned bioacoustic pre-trained models and show the impact of noise-robust pre-training setups. These findings highlight the potential of speech-based self-supervised learning as an efficient framework for advancing bioacoustic research.

\end{abstract}


\section{Introduction}

\Ac{SSL} of speech features has recently led to state-of-the-art performances on linguistic and paralinguistic tasks \cite{superb}. 
\Ac{SSL} methods leverage large volumes of unlabeled data by solving general-purpose pretext tasks and learn contextual and meaningful acoustic representations from the inherent structure of speech. One valid question, given the massive amounts of available speech data, is whether these performances extend to the comparatively under-resourced analysis of non-human animal vocalizations. Recent research has shown that \ac{SSL}\ speech models can reach high performance on a range of tasks such as species recognition \cite{sarkar25}, caller identification \cite{cauzinille24}, or call-type classification \cite{kloots24, sarkar24}. Many such studies have investigated a specific selection of these models with different perspectives, methods, and datasets but lack a comprehensive understanding of how pre-trained speech representations transfer across domains and tasks.

Drawing from previous work, we thus propose a formal study of speech features' transferability:
\begin{itemize}
    \item We benchmark speech models on a set of 10 different bioacoustic tasks, spanning a diversity of taxa including birds, mammals, and insects.
    \item We explore three comparable speech models, showing that robustness to noise and multilingual pre-training are key aspects of out-of-domain transferability.
    \item We compare a set of probing approaches, including linear probing as well as recurrent neural networks, and introduce time-weighted averaging of representations to better leverage their contextual nature.
    \item We confirm the superiority of shallow transformer layers over deeper ones in the context of speech features transferability.
    \item We further analyze the effect of background noise and vocal frequency ranges on a selected dataset. 
\end{itemize} 

\section{Methods}

\begin{figure}[h]
  \centering
  \makebox[\linewidth]{\includegraphics[width=1\linewidth]{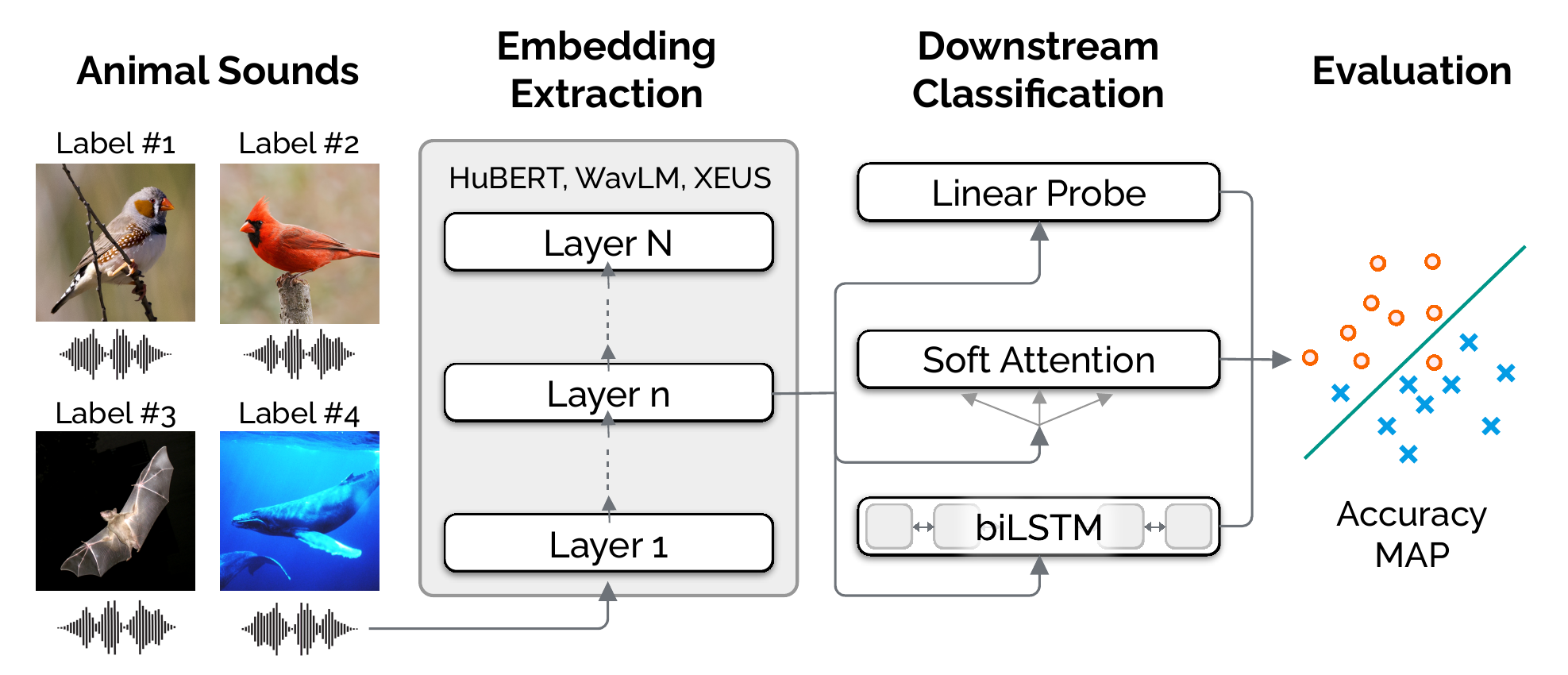}}
  \caption{Workflow of the transfer learning method}
  \label{fig:workflow}
  \vspace{-3pt} 
\end{figure}

This study unfolds as a set of bioacoustic knowledge transfer experiments from three pre-trained \ac{SSL} speech models. Regardless of the selected downstream approach, we initially extract pre-trained representations from each layer of frozen speech models. These representations are subsequently given as train, development, and test sets to a downstream model. Finally, the observed performances are taken as a proxy of the speech model's ability to extract the acoustic information necessary to accomplish the task (see Figure \ref{fig:workflow}). We extend this method, referred to as linear probing \cite{alain2017understanding}, to both non-linear and time-wise analysis of pre-trained representations through other downstream approaches described in Section \ref{sec.downstream}. The code will be made available for replication on GitHub.

\subsection{Tasks and datasets}

The 11 datasets used in this experiment are publicly available as the BEANS benchmark \cite{BEANS}. They include bioacoustic classification and detection tasks on multiple animal taxa, namely ``Watkins'' \cite{watkins} (31 aquatic mammal species classification), CBI \cite{cbi} (264 bird species classification), Egyptian fruit bats \cite{bats} (10 individuals classification), Humbugdb \cite{humbug} (14 mosquito species classification), Dogs \cite{dogs} (10 individuals classification), Dcase \cite{dcase} (20 bird and mammal species detection), Enabirds \cite{enabirds} (34 bird species detection), Hiceas \cite{hiceas} (Minke whale detection), Gibbons \cite{gibbons} (three call-types detection), and RFCX \cite{rfcx} (24 bird and frog species detection). Additionally, ESC-50 consists of classifying 50 environmental and animal sounds. 

For classification, each sample is assigned one or more labels, and evaluation is reported in terms of accuracy. Detection implies identifying subsections of interest and their labels from long recordings with a sliding window approach. Each produced segment may contain multiple labels. Detection evaluation is reported in terms of \ac{mAP}. All datasets are first down-sampled to the required 16 kHz sample rate of pre-trained speech models. They are split into predefined train, development, and test sets with a 6:2:2 ratio, as in \cite{BEANS}. Although all tasks are not directly comparable due to the nature and balance of their labels, dataset sizes, or acoustic environments, they allow a broad overview of speech models' performances in an array of bioacoustic scenarios.

\subsection{Pre-trained models}

\ac{SSL} pre-training consists in leveraging large non-labeled data to pre-train foundation models optimized to extract general-purpose latent representations. With or without fine-tuning, these representations can be given to so-called downstream models, or classification heads, to achieve competitive performance on tasks with small labeled datasets \cite{Mohamed22}. In speech processing, HuBERT \cite{Hubert21}, WavLM \cite{wavlm_22}, and XEUS \cite{XEUS} are \ac{SSL} models originally designed for downstream speech recognition. Additionally, they show high performance on paralinguistic tasks such as speaker identification or emotion recognition, among others \cite{superb}. Both HuBERT and WavLM were successfully implemented in similar bioacoustic experiments \cite{sarkar25, sarkar24, cauzinille24, kloots24}, demonstrating performance that matches or surpasses that of bioacoustic models such as AVES \cite{AVES}, as well as pre-trained bird species classifiers \cite{birdnet} and general-purpose audio taggers. Further details on the three models' architectures are presented in Table \ref{tab:1}.

\begin{table}[h]
    \caption{Pre-trained model specificities}
    \centering
    \resizebox{\columnwidth}{!}{
    \renewcommand{\arraystretch}{1}
    \huge
    \begin{tabular}{ccccc}
        \toprule
        \textbf{Model} & \textbf{Pre-train Data} & \textbf{Arch.} & \textbf{Task} & \textbf{\#param.} \\
        \midrule
        HuBERT large & \makecell{60k hours \\ English only} & 24 TL & MP & 317M \\
        \midrule
        WavLM large & \makecell{94k hours \\ English only} & 24 TL & MP+NR & 90M\\
        \midrule
        XEUS & \makecell{1M+ hours \\ 4k+ languages} & 18 EBL & \makecell{MP+NR\\+Rev} & 577M \\
        \bottomrule
    \end{tabular}}
    \label{tab:1}
    \begin{flushleft}
        \small TL = transformer layers, EBL = E-Branchformer layers, MP = masked prediction, NR = noise robustness, Rev = reverberation robustness.  
    \end{flushleft}
\end{table}

\subsection{Downstream setups}
\label{sec.downstream}

We use linear probing \cite{Ma_2021}, i.e., training a shallow downstream model on a given layer of the \ac{SSL} models, to assess the linear separability of bioacoustic information in their pre-trained latent representation space.
Our first experiment involves training a linear layer on \textbf{Time-Averaged} (T-A) representations as
\begin{align}
\textstyle \bar{x} = \frac{1}{T} \sum_{t=1}^{T} x_t,
\label{(1)}
\end{align}
where \( x_t \in \mathbb{R}^{d} \) is the embedding of the \( t \)-th time frame in the full representation (all model representations are 2-D tensors with embeddings of size 1024 for each 20-millisecond frames). This method allows extracting a single input vector of size 1024 for each example in the dataset and is a preferred option in similar transfer learning experiments \cite{superb}. We hypothesize that averaging latent representations, particularly in such out-of-domain contexts, might end up diluting relevant time-wise information, potentially leading to suboptimal performance for long sound segments.

In a second experiment, we thus introduce \textbf{Time-Weighted Averaging} (T-WA), a form of soft attention that consists of adding a layer of learnable attention weights before averaging all frame representations to better leverage their variability over time. Raw attention scores are obtained as
\begin{align}
\alpha_t = w_\alpha^\top x_t + b_\alpha,
\label{(2)}
\end{align}

where \( w_\alpha \in \mathbb{R}^{d} \) and \( b_\alpha \in \mathbb{R} \) are learned parameters. 
We then compute a weighted average of feature representations as

\begin{align}
\alpha' = \textrm{softmax}_T(\alpha),\\
\textstyle    \tilde{x} = \sum_{t=1}^{T} (\alpha'_t) x_t,
\label{(3)}
\end{align}
where \( \alpha_t \) determines the contribution of each frame. 
This weighted representation is given to a linear layer as in the \ac{T-A} version. The attention weights and the linear probe are trained simultaneously on a single loss. Note that this only adds 1024+1 parameters to the downstream model compared to \ac{T-A}, no matter the input sound length. 

All training procedures were conducted on predefined dataset splits from \cite{BEANS} for 100 epochs. Hyper-parameters such as batch size were experimentally determined for each dataset by comparing metrics on the validation set. We test three learning rates (1e-5, 5e-5, and 1e-4) with an Adam optimizer and weight decay for each experiment and display the best observed result for the best layer. Classification tasks are conducted with cross entropy loss and evaluated in terms of accuracy. Multi-label detection tasks are conducted with BCE loss and evaluated in terms of \ac{mAP}. Random baselines are computed by selecting the highest possible score between predictions of only the most frequent label (for highly unbalanced datasets) or random predictions.

\section{Results}

Table \ref{tab:results} shows accuracy and \ac{mAP} of the best performing layer on the validation set for each dataset, model, and linear probing approach, as well as the results from fully fine-tuned AVES bio \cite{AVES}, a Vggish model, and the best CNN-based approach from the BEANs benchmark \cite{BEANS} for comparison. Underlined results inform on performance gains between benchmark results and our solutions.

\begin{table*}[h]
    \caption{Probing Results}
    \centering
    \begin{tabular}{|c|c|c|c|c|c|c|c|c|c|c|c|}
        \cline{2-12}
         \multicolumn{1}{c|}{} & \multicolumn{5}{c|}{accuracy} & \multicolumn{5}{c|}{\ac{mAP}} & accuracy \\ \hline
         Datasets & watkins & cbi & bats & humbug & dogs & dcase & enabirds & hiceas & gibbons & rfcx & esc-50\\ \hline\hline
         HuBERT T-A & 0.89 & 0.28 & 0.62 & 0.76 & 0.79 & 0.32 & 0.48 & 0.54 & 0.19 & \underline{\textbf{0.14}} & 0.72 \\ \hline
         WavLM T-A & \underline{\textbf{0.90}} & 0.34 & 0.68 & 0.79 & 0.84 & 0.33 & \textbf{0.51} & 0.54 & 0.18 & 0.10 & 0.76\\ \hline
         XEUS T-A & 0.89 & 0.25 & 0.73 & \textbf{0.79} & 0.86 & 0.30 & 0.44 & 0.54 & 0.26 & 0.08 & \underline{\textbf{0.81}}\\ \hline \hline
         HuBERT T-WA & 0.85 & 0.37 & 0.74 & 0.75 & 0.90 & 0.33 & 0.44 & 0.58 & 0.23 & 0.13 & 0.57\\ \hline
         WavLM T-WA & 0.84 & \textbf{0.38} & 0.76 & 0.74 & 0.93 & \underline{\textbf{0.39}} & 0.44 & \underline{\textbf{0.64}} & \underline{\textbf{0.32}} & 0.12 & 0.65\\ \hline
         XEUS T-WA & 0.78 & 0.28 & \underline{\textbf{0.77}} & 0.75 & \underline{\textbf{0.95}} & 0.32 & 0.47 & 0.52 & 0.28 & 0.10 & 0.68\\ \hline \hline
         AVES bio ft & 0.88 & \underline{0.60} & 0.75 & \underline{0.81} & \underline{0.95} & \underline{0.39} & \underline{0.56} & 0.63 & 0.28 & 0.13 & 0.77\\ \hline
         Vggish & 0.85 & 0.44 & 0.74 & \underline{0.81} & 0.91 & 0.34 & 0.54 & 0.46 & 0.15 & \underline{0.14} & 0.77\\ \hline
         CNN (best) & 0.80 & 0.57 & 0.74 & 0.70 & 0.76 & 0.22 & 0.46 & 0.30 & 0.31 & 0.09 & 0.59\\ \hline
         random & 0.07  & 0.01  & 0.10 & 0.35  & 0.22  & 0.03  & 0.48  & 0.27  & 0.02 & 0.01 & 0.02 \\ \hline
    \end{tabular}
    \begin{center}
        \small (ft: fine-tuning) 
        Best speech model's layer performance is in \textbf{bold}, best performance compared with benchmark is \underline{underlined}. 
    \end{center}
    \label{tab:results}
\end{table*}

The results show the intrinsic ability of pre-trained speech representations to encode enough information to resolve bioacoustic tasks across taxa with linear probing. All models show variable results across layers. Depending on the dataset, performances increase in the initial layers, reach a maximum between layer 3 and 11 for HuBERT, layer 4 and 15 for WavLM, and layer 2 and 6 for XEUS, then consistently drop in deeper layers. This behavior was previously observed in similar experiments \cite{sarkar24,sarkar25,cauzinille24} and stays consistent with speech models layer-wise acoustic information encoding as described in \cite{pasad23}.
Both WavLM and XEUS show higher probing performance than HuBERT, which we discuss in Section \ref{sec:models}. \Ac{T-WA} representations marginally improve results compared to \ac{T-A} ones. This shows that preserving time-wise information may be significant for datasets with longer sound samples, as discussed in Section \ref{sec:time}. 

In such cross-species transfer learning experiments, one could make the hypothesis that the phylogenetic proximity of a species with humans results in similarities between the vocalizations of said species and speech. This can be based on the resemblance of vocal tracts, anatomical sizes, vocal frequencies, degrees of sequentiality, etc. Although the tasks presented here are difficult to compare due to differences in dataset sizes, annotations, label distribution, acoustic environments, and sound lengths, there seems to be no effect of species proximity with humans on the observed performance. Pre-trained speech models show a general ability to encode bioacoustic information regardless of the species. Furthermore, results on the esc-50 dataset tend to show speech models' ability to work as general-purpose audio taggers. We thus extend this experiment to account for the effect of time, noise, and pre-training setup in an attempt to better understand the mechanisms underlying such results. 

\section{Discussion and perspectives}

We identify four factors that may account for performance variations between models and datasets. The effect of background noise is discussed in Sections \ref{sec:models}, and \ref{sec:freq}, species vocalization overlap is investigated in Section \ref{sec:models}, time-related factors in Section \ref{sec:time} and frequency range mismatch with speech in Section \ref{sec:freq}.

\subsection{Model comparison}
\label{sec:models}

WavLM and XEUS include noise robustness solutions by being pre-trained with dynamic mixing of artificial noise and speech overlap (as well as simulated reverberant conditions for XEUS). This improves resilience to background noise, a significant limiting factor in bioacoustic tasks, since such recordings typically have much lower sound-to-noise ratios than speech datasets. Although all three models show additional differences in their pre-training data (see Table \ref{tab:1}), we argue that the noise robustness of WavLM and XEUS plays an important role in their higher performances. WavLM, in particular, with significantly fewer parameters, outperforms both other models in six out of the 11 tasks. 

Similarly to background noise, overlapping animal vocalizations may also harm performance in some datasets. For enabirds and rfcx, the observed low performances, either in terms of relative performance to the random baseline (enabirds) or regarding generally low scores (rfcx), can be linked to a significant amount of interspecies vocalization overlap present in both datasets. The enabirds dataset is comprised of recordings of bird choruses, which can contain over six species labels in a single sound sample. Similarly, rfcx recordings are drawn from tropical forest environments with high multi-species presence. In both cases, WavLM, and XEUS pre-training strategy involving artificial mixup of two speech utterances does not seem to result in sufficient robustness to overcome the difficulties in disentangling simultaneous vocalizing species. Conversely, pre-trained bioacoustic models such as Perch \cite{perch} and Birdnet \cite{birdnet} have been shown to increase multi-label classification performance with multi-species mixup training.

The multilingual pre-training of XEUS may also explain some of the performance improvements in 4 out of 11 tasks, as English-only pre-training datasets are inherently less acoustically variable than multilingual ones. However, this assumption would need further investigation, as XEUS is larger and trained on significantly more data compared to both other models. 

\subsection{Time information}
\label{sec:time}

Excluding enabirds and rfcx, two datasets showing relatively low performance in general, \ac{T-A} representations consistently outperform \ac{T-WA} on datasets with sound segments shorter than 3.98 seconds. \ac{T-A} representations do not entirely remove time-wise information, as each frame representation still contains contextual information from previous and future frames, allowing good performance on short sound segments despite averaging. Yet, \ac{T-WA} results indicate that relying on a learned selection of relevant sound frames according to their representation is an important feature for datasets containing longer sound segments.

To further analyze the effect of temporal information and linear separability on the observed performances, we replace linear probes with Echo State Networks (ESN) \cite{sun20} or Bidirectional Long Short-Term Memory networks (biLSTM). Both can extract non-linear information from entire representations without risking a loss of time-sensitive information. ESNs, derived from reservoir computing, mostly consist of an RNN with randomly initialized fixed weights. They are a good approximation of a recurrent non-linear solution that does not demand learning additional parameters compared to linear probing. BiLSTMs, on the other hand, significantly augment the number of parameters but should produce a strong topline solution with sufficient amounts of training data.

Interestingly, ESNs always showed worse or on par performance compared to linear probing on all datasets. BiLSTMs showed some improvements for HuBERT humbug (0.77 accuracy with a two-layer biLSTM), HuBERT gibbons (0.26 mAP with a single-layer biLSTM), XEUS enabirds (0.5 mAP with a single-layer biLSTM) and XEUS gibbons (0.35 mAP with a single-layer biLSTM). Except for this outlying last case, biLSTMs always underperform linear probes from either XEUS or WavLM.
In conclusion, linear probes applied to time-averaged or time-weighted latent representations seem to outperform more complex recurrent models in bioacoustic tasks, as they may strike a better balance between signal extraction and overfitting. 

\subsection{Frequency shift and noise addition}
\label{sec:freq}

\begin{figure}
    \centering
    \includegraphics[width=\linewidth]{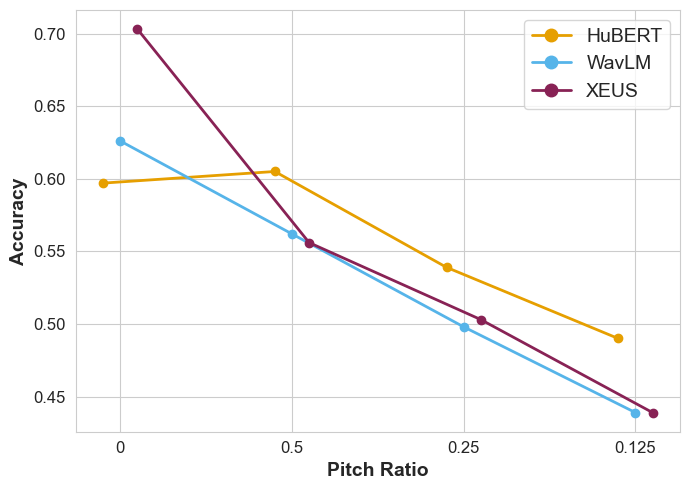}
    \caption{Performance for the Egyptian fruit bats dataset on the 10th layer with pitch shifting (T-A)}
    \label{fig:pitch}
\end{figure}

From elephant rumbles to bat calls, many animal vocalizations sit below or above the human speech range. We design an ablation experiment to account for the effects of frequency range on the Egyptian fruit bats dataset. 
Pre-trained speech models are trained on relatively clean recordings of speech at 16 kHz, thus ranging from 0 to 8,000 Hz, with most of the energy between 90 and 2,500 Hz. 
This could lead to some degree of overfitting to clean signals centered around this particular frequency range, hindering performance on higher-pitched or noisy signals. 
We investigate whether pitch downshifting of ultrasonic bat calls results in improved performance by reducing their frequency towards the speech range. We progressively lower the pitch of the recordings at 16kHz with factors of 0.125, 0.25, and 0.5 by resampling, then interpolating them with the \texttt{rate} function from \texttt{pysox}. Surprisingly, Figure \ref{fig:pitch} shows that this lowers the probe's accuracy, regardless of the model's layer used. We also observed this tendency in similar experiments not included in this manuscript: time-stretching as well as pitch-shifting the original 250 kHz samples. In these cases, adding high-frequency information from ranges above 8 kHz did not improve performance either.

These results show that speech models may be robust to high frequencies, potentially through their modeling of fricatives. Additionally, pitch shifting as a transformation involves resampling and interpolation, which may result in unnatural audio quality when performed at such extreme ratios. 

\begin{figure}
    \centering
    \includegraphics[width=\linewidth]{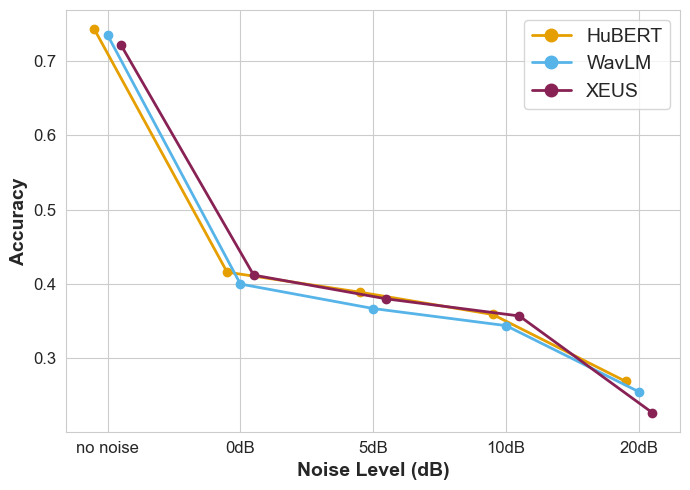}
    \caption{Performance for the Egyptian fruit bats dataset on the 10th layer with noise addition (T-A)}
    \label{fig:noise}
\end{figure}

We also explore the effect of noise on classification accuracy by mixing natural noise from various sources \cite{fonseca2021fsd50k,salamon2014dataset,thiemann2013diverse,abesser2021idmt,wichern2019wham} with the bats vocalizations at \acp{SNR} of 0, -5, -10, and -20 dB. Note that these are extreme factors for speech applications but quite common in bioacoustics. As seen in Figure \ref{fig:noise}, the accuracy decreases with the \ac{SNR}. Interestingly, performance stays above random even at extreme \acp{SNR} such as -20dB, possibly because the added environmental noise does not fully mask the high frequencies of bats calls.
\looseness=-1

Both these results indicate some degree of robustness of speech models to bioacoustic transfer learning while partly explaining the variability of their performance across scenarios. Low performance on datasets such as cbi, gibbons, rfcx and enabirds could be explained both by the difficulty of the task or label distribution and because of the significantly low \acp{SNR} of the recordings. \\

\subsection{Comparing pre-trained models with random initializations}

\label{sec:random}

To reach a better understanding of the effect of pre-training on the observed performances of speech models, we test the intuitive hypothesis stating that randomly initialized models should show results on par with random baselines (as only the pre-training phase should result in model parameters able to capture acoustic information). Interestingly, a randomly initialized HuBERT model achieves slightly better-than-chance performance on bioacoustic tasks when evaluated using linear probing, with an average accuracy increase of about 10\%. While the model lacks any learned knowledge of speech or acoustic structure, this result aligns with a growing body of evidence suggesting that untrained deep learning models can provide non-trivial representations due to their inherent inductive biases. It also shows that the discriminative capabilities of SSL models primarily stem from their speech-based pretraining phase, although a small fraction of their effectiveness on bioacoustic tasks may be attributed to these structural priors \cite{Rahimi_2008, Ulyanov_2018}.

\subsection{Limitations}
\label{sec.limits}

Comparing performance across datasets must be taken with care due to differences in dataset sizes, tasks, label distribution, etc. We limit our conclusions to relative improvements of specific methods rather than absolute performance differences. Additionally, the use of multiple random seeds would allow getting a stronger sense of performance gaps between models. Here, each training was performed once, from fixed dataset splits provided in the BEANS benchmark. 
We also advocate for a cautious comparison of our results with AVES-bio, which is based on a significantly smaller model (HuBERT base), pre-trained with in-domain data, and fully fine-tuned on each task during training. Although our experiments are not performance-oriented, we must mention that fine-tuning of large bioacoustic foundation models with comparable sizes generally outperforms our methods on most of the benchmark \cite{robinson23, robinson24, Deng_2025}. Conversely, task-specific machine learning solutions and CNNs trained on Mel Frequency Cepstral Coefficient representations are mostly outperformed by frozen speech-based probing \cite{BEANS}.

\section{Conclusion}

This study shows the effective knowledge transfer capabilities of self-supervised speech models for bioacoustic tasks. In particular, HuBERT, WavLM, and XEUS generate rich representations, resulting in competitive performance on classification and detection tasks across taxa.
Our findings indicate that phylogenetic proximity to humans does not influence this transfer, emphasizing the generalizability of speech representations in out-of-domain scenarios. Variability in results across datasets highlights the impact of task complexity and acoustic environments. Notably, the presence of noise greatly impacts linear probing results, and noise-robust pre-training is advantageous.
Temporal information remains crucial, with time-weighted averaging approaches outperforming time-averaging of representations for longer audio samples. However, recurrent networks showed scarce benefits compared to linear probing, likely due to overfitting issues. Finally, speech models exhibit some degree of robustness to the frequency range of bioacoustic signals, even with high-pitched vocalizations. Overall, self-supervised speech models show strong promise for the development of foundation models in bioacoustics, offering interesting solutions related to data quality and quantity. Leveraging large speech datasets and SSL architectures, future work should focus on developing bioacoustic models partly pre-trained on speech with enhanced noise robustness and improved mixup strategies.

\section{Acknowledgments}

This work, carried out within the Institute of Convergence ILCB, was supported by grants from France 2030 (ANR-16-CONV-0002), the Excellence Initiative of Aix-Marseille University (A*MIDEX), the HEBBIAN ANR project (\#ANR-23-CE28-0008), and was partially funded by the Earth Species Project.

\bibliographystyle{IEEEtran}
\bibliography{refs}

\end{document}